\documentclass{article} 
\usepackage{iclr2027_conference,times}

\usepackage{amsmath,amsfonts,bm}

\def\eqref#1{equation~\ref{#1}}

\def\1{\bm{1}}

\DeclareMathAlphabet{\mathsfit}{\encodingdefault}{\sfdefault}{m}{sl}
\SetMathAlphabet{\mathsfit}{bold}{\encodingdefault}{\sfdefault}{bx}{n}

\DeclareMathOperator*{\argmin}{arg\,min}

\usepackage{hyperref}
\usepackage{url}

\usepackage{amsmath}
\usepackage{booktabs}
\usepackage{colortbl}
\usepackage{multirow}
\usepackage{graphicx}
\usepackage{tablefootnote}
\usepackage[normalem]{ulem}
\usepackage{wrapfig}

\newcommand{\ours}{ProFiT~}
\newcommand{\ourscomma}{ProFiT}
\newcommand{\std}[1]{\ensuremath{{}_{\scriptscriptstyle #1}}}

\title{Flow-Matching-Based Protein Structure Tokenizer Made Efficient and Easy}

\author{Zhe Zhang$^{1,2}$ \thanks{Zhe Zhang (zhe-zhan22@mails.tsinghua.edu.cn).} \quad Yikai Zhang$^{3}$ \quad Jiangtao Feng$^{1}$ \quad Ya-Qin Zhang$^{1}$ \quad Wei-Ying Ma$^{1}$ \quad Hao Zhou$^{1}$ \thanks{Correspondence to Hao Zhou
(zhouhao@air.tsinghua.edu.cn).} \\
$^1$ Institute for AI Industry Research (AIR), Tsinghua University \\
$^2$ Department of Computer Science and Technology, Tsinghua University \\
$^3$ Fudan University
}

\iclrfinalcopy 
\begin{document}

\maketitle

\begin{abstract}
As the bridge between protein modality and discrete modeling, protein structure tokenization still largely relies on heavily engineered training objectives tailored to specific downstream tasks and large training datasets, which hinders its transfer to broader application scenarios. To address this issue, we propose \ourscomma, a lightweight flow matching tokenizer. With simple training strategies that encourage healthy codebook utilization, \ours can be trained efficiently and naturally learns semantically meaningful representations without any manual semantic alignment, while achieving reconstruction quality and generalization that match or surpass those of substantially larger tokenizers. We conduct extensive evaluations across a wide range of settings and demonstrate that \ours is a plug-and-play tokenizer adaptable to diverse downstream tasks. This study further reveals the significant potential of the flow matching tokenizer paradigm. Our code is publicly available at \url{https://github.com/QDKStorm/ProFiT}.
\end{abstract}

\section{Introduction}
The goal of computational biology modeling is shifting from task-specific optimization toward general modeling of broad biological capabilities \citep{elnaggar2021prottrans,nijkamp2023progen2,heinzinger2024bilingual,wang2025dplm,hayes2025simulating,ma2025prottex}. Many such general-purpose models benefit from the discretization of biological data, particularly proteins. Once discretized, protein representations can be jointly modeled with other modalities (e.g., text and amino acid sequences) and naturally align with autoregressive and masked language modeling (MLM) frameworks, thereby benefiting from the scaling laws of Transformers and discrete modeling paradigms \citep{nijkamp2023progen2,wang2025dplm,zhang2024balancing,dilip2026adaptive}. Therefore, protein structure tokenization serves as a bridge between continuous 3D geometry and discrete generative modeling.

Recent years have seen considerable exploration of protein structure tokenization. Starting with Foldseek, researchers began investigating the discretization of protein structures \citep{van2024fast}. Subsequent studies have progressively explored and advanced their performance in reconstruction and certain downstream tasks. Much of this work builds on classical $SE(3)$-equivariant architectures and training methods, among which some tokenizers have demonstrated the ability to generate proteins under specific architectures \citep{lin2023protokens,wang2025dplm,zhang2024balancing,gaujac2024learning,hayes2025simulating}. Later studies further improved aspects such as codebook utilization and training stability \citep{gao2024foldtoken2,gao2024foldtoken3,gao2024foldtoken4,gao2025foldtoken,yuan2025protein}.

However, existing tokenizers face several limitations:
(i) To balance reconstruction quality and semantic preservation, training objectives are highly engineered, requiring careful weighting of multiple losses for geometric alignment, confidence, and semantic consistency.
(ii) Most architectures rely on single-step inference, which severely constrains reconstruction fidelity and leaves little flexibility for simultaneously optimizing reconstruction and semantic representation.
(iii) Tokenizers are typically trained on real structures and synthetic data generated by folding models, introducing bias in the output distribution. As a result, tokenizers tend to produce plausible structures rather than faithfully reflecting the encoded input \citep{lin2023evolutionary,fleming2025alphafold}.
(iv) Most practical tokenizers, such as DPLM2 and ESM3, rely on pretraining over large-scale structural datasets, making it prohibitively expensive to train a tokenizer from scratch with comparable~\citep{wang2025dplm,hayes2025simulating}.

In the image domain, flow-matching-based tokenizers have begun to emerge and have been shown to address the aforementioned limitations \citep{lipman2022flow,chen2025diffusion,sargent2025flow}. These works demonstrate that the flow matching objective alone is sufficient to train high-performance tokenizers, and further suggest that learning the velocity field naturally induces generalization and semantically meaningful representations. Moreover, due to its simple and unified loss formulation, flow matching exhibits strong scalability to larger datasets and more efficient parameter utilization \citep{chen2025diffusion,sargent2025flow}.

We argue that although these phenomena have been preliminarily validated in protein tokenizers, they are still far from being fully explored \citep{dilip2026flow,dilip2026adaptive}. Previous work has only validated the feasibility of flow-matching–based tokenization on relatively small-scale data, lacking both an explanation of how semantic representations emerge during flow matching and effective strategies to encourage this phenomenon. In addition, there is still a lack of unified evaluations spanning diverse scenarios and tasks to examine whether protein structure tokenizers possess generalizable capabilities. Therefore, we propose \ourscomma, which more fully exploits the potential of flow matching and provides a parallel comparison between \ours and classical tokenizers across a broad range of settings. Our main contributions are summarized as follows:

\begin{enumerate}
    \item We propose \ourscomma, a protein structure tokenizer that establishes a new state of the art among flow-matching-based approaches using a simple and effective training recipe.
    \item We systematically investigate how to stabilize flow matching training on protein data, as well as techniques that encourage stronger semantic representations and better generalization during training.
    \item We conduct extensive evaluations of \ours in terms of reconstruction quality and semantic understanding, as well as its performance across diverse generative model architectures and tasks, including both unconditional and conditional generation. To our knowledge, this is the first systematic study of the adaptability of multiple protein structure tokenizers specifically across different generative architectures and tasks.
\end{enumerate}

\begin{figure}[t]
    \centering
    \includegraphics[width=1\linewidth]{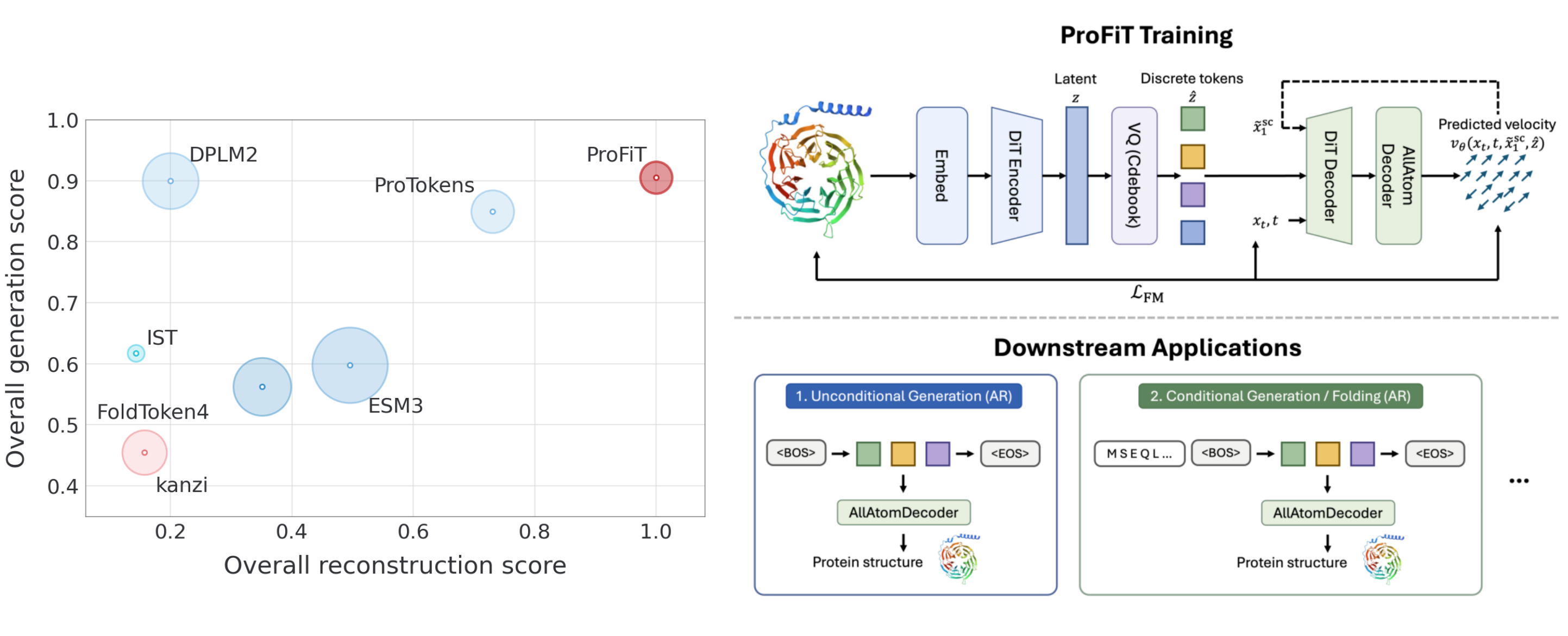}
    \caption{(Left) Comparison between tokenizer scale and performance. The horizontal and vertical axes denote the overall reconstruction and generation scores (defined in App.\ref{app:overall_score_definitions}). The radius of each point is proportional to the logarithmically scaled tokenizer parameter count. Flow-matching–based tokenizers are highlighted in \textcolor{red}{red}, while other tokenizers are shown in \textcolor{blue}{blue}. All figures in this paper follow this visualization convention. (Right) the training framework of \ours and its supported downstream applications.}
    \label{fig:main}
\end{figure}

\begin{figure}[t]
    \centering
    \includegraphics[width=1\linewidth]{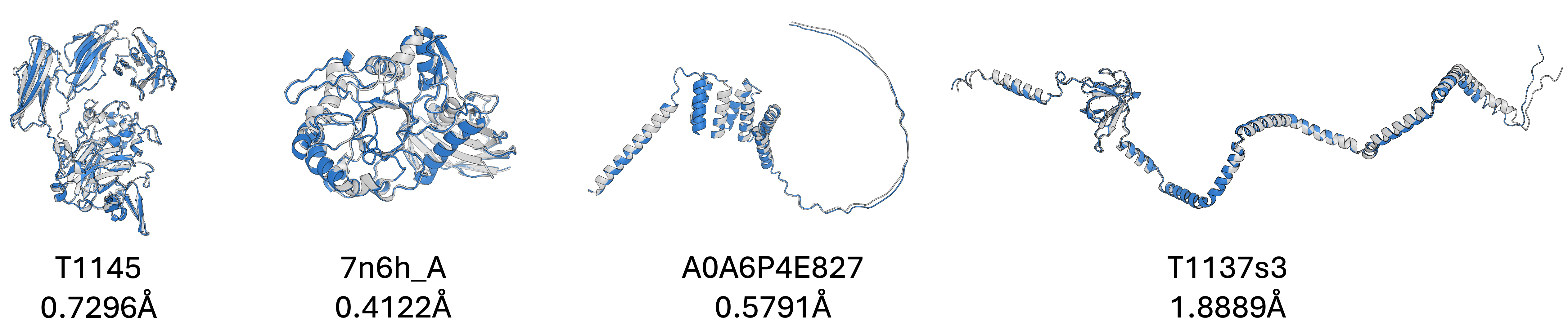}
    \caption{protein reconstruction examples of \ourscomma. \textcolor{gray}{Gray}: ground truth, \textcolor{blue}{blue}: \ourscomma.}
    \label{fig:examples}
\end{figure}

\section{Related Work}
\subsection{Diffusion Model}
Diffusion models formulate generation as gradually transforming a simple noise distribution into data through a time-dependent stochastic or deterministic process. Early denoising diffusion probabilistic models (DDPMs) established the forward noising and reverse denoising framework, while score-based methods provided a continuous-time view via SDEs and linked diffusion training to score estimation \citep{ho2020denoising,song2020score}. Subsequent work improved efficiency and sample quality through accelerated samplers (e.g., non-Markovian/ODE-style sampling), improved parameterizations, and latent-space diffusion that shifts generation from pixel space to compact representations \citep{song2020denoising,nichol2021improved,lu2022dpm,liu2022pseudo,karras2022elucidating,rombach2022high}. More recently, probability-flow perspectives unified diffusion and transport-based generative modeling: rectified flow reframes training around straightened trajectories, and flow matching directly regresses the target velocity field along prescribed probability paths \citep{lipman2022flow,liu2022rectified}. Compared with classical diffusion objectives, flow matching offers a simpler single-loss formulation, flexible path design, and strong scalability, making it an increasingly practical foundation for large-scale generative modeling, including scientific and biological domains \citep{esser2024scaling}.

\subsection{Protein Structure Tokenization}
Protein structure tokenization aims to compress continuous 3D protein structure coordinates into a discrete token sequence, thereby adapting it for downstream tasks. Early attempts at discretizing the protein modality, such as Foldseek, introduced highly compact structural alphabets that capture residue-level geometric interactions for efficient retrieval; however, the strong information bottleneck prevents them from performing well on reconstruction tasks \citep{van2024fast}. The dominant paradigm has since evolved toward discrete autoencoder frameworks, where an encoder maps coordinates ($\mathrm{C}_\alpha$, backbone, or all-atom) into latent representations, a quantizer discretizes them into tokens, and a deterministic decoder reconstructs 3D structures. Representative models differ primarily in quantization schemes, vocabulary sizes, encoder equivariance, and training objectives \citep{lin2023protokens,wang2025dplm,zhang2024balancing,gao2024foldtoken2,gao2024foldtoken3,gao2024foldtoken4,hayes2025simulating,gao2025foldtoken}. More recently, flow-based autoencoders replace deterministic decoders with conditional flow matching, modeling structure generation as a continuous denoising trajectory from noise to coordinates \citep{dilip2026flow,dilip2026adaptive}. This formulation simplifies training objectives into a single flow loss while maintaining competitive reconstruction quality and significantly reducing model and data requirements. Despite rapid progress, the design space of protein structure tokenizers remains highly unconstrained, with limited consensus on architecture or quantization strategies. Moreover, most tokenizers can maintain strong reconstruction ability, but only some tokenizers demonstrate general adaptability to generative models.

\section{Preliminaries}
\subsection{Flow Matching}
Flow matching provides a framework for learning continuous-time generative models by directly regressing a velocity field that transports a simple base distribution to a target data distribution~\citep{lipman2022flow}. Let $p_0$ denote a tractable base distribution and $p_1$ the data distribution. The goal is to construct a time-dependent probability path $\{p_t\}_{t \in [0,1]}$ connecting $p_0$ to $p_1$, and learn a velocity field $v_\theta(x, t)$ such that samples evolve according to the ordinary differential equation (ODE)
\begin{equation}
\frac{d x_t}{d t} = v_\theta(x_t, t), \quad x_0 \sim p_0,
\end{equation}
and the marginal distribution of $x_t$ matches $p_t$ for all $t \in [0,1]$.

Flow matching avoids explicit likelihood computation by supervising the model with a target velocity field derived from a prescribed conditional probability path. Specifically, given a coupling between $x_0 \sim p_0$ and $x_1 \sim p_1$ and the path $x_t = \phi_t(x_0, x_1)$, the ground-truth velocity field is then given by
\begin{equation}
v^*(x_t, t) = \frac{d}{dt} \phi_t(x_0, x_1),
\end{equation}
which induces a marginal flow consistent with $p_t$. The model is trained by minimizing
\begin{equation}
\mathcal{L}_{\text{FM}} = \mathbb{E}_{t, \, (x_0, x_1)} 
\left[ \| v_\theta(x_t, t) - v^*(x_t, t) \|_2^2 \right].
\end{equation}
This formulation decouples the choice of probability path from the model parameterization, allowing flexible design of interpolation schemes. A common choice is $x_0 \sim \mathcal{N}(0, I), x_1 \sim p_{\text{data}}$, and defines the conditional path $x_t = \alpha(t) x_1 + \sigma(t) x_0$, where $\alpha(t)$ and $\sigma(t)$ are scalar schedules satisfying boundary conditions $\alpha(0)=0, \alpha(1)=1$ and $\sigma(0)=1, \sigma(1)=0$. This construction ensures $x_0 \sim p_0$ and $x_1 \sim p_1$, while interpolating between noise and data. The corresponding target velocity field is obtained by differentiating the path:
\begin{equation}
v^*(x_t, t) = \dot{\alpha}(t) x_1 + \dot{\sigma}(t) x_0.
\end{equation}

\section{Method}
\label{sec:method}
We cast protein backbone tokenization as a flow-matching autoencoder with a discrete bottleneck. Given a protein of length $L$, let its backbone coordinates be $\mathbf{X} = \{\mathbf{x}_i\}_{i=1}^{L}, \mathbf{x}_i \in \mathbb{R}^{4\times 3}$, where each residue contains the heavy atoms $\{\mathrm{N}, \mathrm{C}_\alpha, \mathrm{C}, \mathrm{O}\}$ (measured in nanometers). Our objective is to learn a mapping $\mathbf{X} \mapsto \mathbf{c} \in \{1,\dots,K\}^{L}$ that compresses a continuous structure into a sequence of discrete tokens while preserving faithful reconstruction, robustness to perturbations, and high codebook utilization.

\subsection{Overall Architecture}
The architecture of \ours follows prior practices for modeling proteins with non-equivariant networks~\citep{abramson2024accurate,li2026proteinae}. The encoder maps an input backbone structure $\mathbf{x}_1 \in \mathbb{R}^{L \times 4 \times 3}$ to a latent representation $\mathbf{z}$ through $N_{\mathrm{enc}}$ DiT blocks with pair bias:
\begin{equation}
    \begin{aligned}
    \mathbf{s}_0 = \mathrm{Embed}(\mathbf{x}_1), \quad
    \mathbf{s}_i = \mathrm{DiT}_i(\mathbf{s}_{i-1}, \mathbf{x}_1), i = 1,\dots,N_{\mathrm{enc}}, \quad
    \mathbf{z} = \mathrm{Linear}(\mathbf{s}_{N_{\mathrm{enc}}}) \in \mathbb{R}^{L \times d}.
    \end{aligned}
\end{equation}
We use an Atom Transformer as the $\mathrm{Embed}$ layer~\citep{abramson2024accurate}. Since \ours operates only at the backbone level, it aggregates backbone atom features into residue-level representations. The pair bias in each encoder DiT block is constructed from binned inter-residue $\mathrm{C}_\alpha$ distances and sequence indices of $\mathbf{x}_1$. The final token-level representation $\mathbf{s}_{N_{\mathrm{enc}}}$ is projected into an extremely low-dimensional space, forming a tight information bottleneck.

The continuous latent $\mathbf{z}$ is discretized by a vector quantizer $\mathcal{Q}$ with codebook $\mathcal{E} = \{\mathbf{e}_k\}_{k=1}^{K} \subset \mathbb{R}^{d}$, $K = 8192$~\citep{van2017neural}. For each residue position $i$,
\begin{equation}
    c_i = \argmin_{k \in [K]} \lVert \mathbf{z}_i - \mathbf{e}_k \rVert_2^{2},
    \qquad
    \hat{\mathbf{z}}_i = \mathbf{e}_{c_i},
\end{equation}
with gradients propagated through a straight-through estimator.

The decoder reconstructs the structure from a noisy input $\mathbf{x}_t$ at flow-matching timestep $t$, conditioned on the quantized latent $\hat{\mathbf{Z}}$ and using the same DiT stack but with different conditioning signals:
\begin{equation}
    v_\theta(\mathbf{x}_t, t, \hat{\mathbf{Z}}) = \mathrm{AllAtomDecoder}\left(\mathrm{DiTStack}\left(\mathbf{x}_t, \tilde{\mathbf{x}}_1^{\mathrm{sc}}, t, \hat{\mathbf{Z}}\right)\right),
\end{equation}
The decoder’s pair bias is constructed from binned pairwise distances derived from both the noisy trajectory $\mathbf{x}_t$ and the self-conditioned previous-step prediction $\tilde{\mathbf{x}}_1^{\mathrm{sc}}$, along with sequence indices and a time embedding. $\mathrm{AllAtomDecoder}$ module directly outputs the velocity field $v_\theta(\mathbf{x}_t, t, \hat{\mathbf{Z}})$ used in flow matching. The discrete tokens thus serve as a semantic token of the structure, while geometric detail is reconstructed by a conditional continuous-time generative model.

\subsection{Enhancements to Vector Quantization}
\label{sec:enhancements_VQ}
Vanilla VQ training is notoriously known to lead to weaker model performance and an imbalanced codebook distribution. We observe two types of phenomena that degrade the performance of \ourscomma: (i) \textbf{codebook collapse}, including but not limited to low codebook utilization and the tendency of code norms to approach zero during training and eventually become unused; and (ii) \textbf{decoder overfitting}, where the downstream decoder overfits to the exact distribution of $\hat{\mathbf{Z}}$ and degrades sharply when the token sequence is even mildly perturbed. We address these issues with two complementary mechanisms.

\paragraph{Random Token Masking.}
At inference time, token sequences produced by a downstream prior inevitably deviate from the ground-truth encoding; a decoder trained only on clean conditioning will overfit and degrade rapidly. To build error tolerance, at each training step we sample a corruption ratio $r' \sim \mathcal{U}(0, r)$ where $r$ is a fixed hyperparameter, then independently mask each quantized token $\hat{\mathbf{z}}_i$ with probability $r'$:
\begin{equation}
    \tilde{\mathbf{z}}_i = (1 - m_i)\hat{\mathbf{z}}_i, \qquad m_i \sim \mathrm{Bern}(r'), \qquad i = 1,\dots,L.
    \label{eq:rtm}
\end{equation}

\paragraph{Dead Code Replacement.}
In our training, we observe that among randomly initialized codes, a large proportion tend to be unused. In EMA-based VQ updating, these unused codes receive updates only through EMA decay, causing their norms to gradually converge to zero and eventually become dead codes, which undermines efficient codebook utilization. The underlying causes of this phenomenon may be multifaceted \citep{fifty2025restructuring}. A choice is to maintain an EMA cluster size $C_k^{(s)}$ for each code $k$ with decay $\gamma$, updated as $C_k^{(s)} = \gamma C_k^{(s-1)} + (1-\gamma) \sum_{i: c_i = k} 1$ at each step \citep{zheng2023online}. We additionally, after each update, replace any code whose $C_k^{(s)}$ falls below a threshold $\tau_{\text{dead}}$ by a vector sampled uniformly from the current batch's encoder outputs, and its EMA statistics are reset to a predefined mass $\rho > \tau_{\text{dead}}$. This reset mass acts as a grace period that prevents the revived code from being immediately declared dead again, giving it time to attract real assignments.

\subsection{Training Objective}
\label{sec:objective}

Let $\pi_{\text{data}}$ denote the distribution over backbone coordinates after centroid removal and random-rotation augmentation. Following rectified flow~\citep{liu2022rectified} we construct the linear probability path $\mathbf{x}_t = (1-t)\mathbf{x}_0 + t\mathbf{x}_1,\mathbf{x}_0 \sim \mathcal{N}(\mathbf{0}, \mathbf{I}_{3}),\mathbf{x}_1 \sim \pi_{\text{data}},t \in [0,1]$, whose target velocity is
\begin{equation}
    \mathbf{v}^{\star}_t = \frac{\mathbf{x}_1 - \mathbf{x}_t}{1-t}.
\end{equation}
The decoder directly predicts the velocity field $v_\theta(\mathbf{x}_t, t, \tilde{\mathbf{Z}})$. With $\tilde{\mathbf{Z}}$ obtained from Eq.~\eqref{eq:rtm}, we use the flow-matching loss over all $4L$ backbone atoms as the training objective,
\begin{equation}
    \mathcal{L}_{\text{FM}}(\phi,\theta)
    =
    \mathbb{E}_{\mathbf{x}_1,\mathbf{x}_0,t}\left[
    \frac{1}{12L}\sum_{i=1}^{4L}\left\lVert \mathbf{v}^{\star}_{t,i} - v_{\theta,i}(\mathbf{x}_t, t, \tilde{\mathbf{Z}}) \right\rVert_{2}^{2}
    \right].
    \label{eq:fm_loss}
\end{equation}
We use the same t-sampling distribution as ProteinAE: $t \sim 0.2\cdot \mathcal{U}(0,1) + 0.8\cdot\mathrm{Beta}(1.9,1.0)$ \citep{li2026proteinae}.

\subsection{Sampling}
During inference, we start from $\mathbf{x}_0 \sim \mathcal{N}(\mathbf{0}, \mathbf{I}_{3})$ and the codebook embeddings $\hat{\mathbf{Z}} = \mathcal{Q}^{-1}(\mathbf{c})$ without masks, we sample from the unified SDE below, whose ODE form is recovered by setting $\eta = \gamma = 0$:
\begin{equation}
    \mathrm{d}\mathbf{x}_t
    =
    \left[v_\theta(\mathbf{x}_t,t,\hat{\mathbf{Z}}) + g(t)\eta s_\theta(\mathbf{x}_t,t,\hat{\mathbf{Z}})\right]\mathrm{d}t
    +
    \sqrt{2g(t)\gamma}\mathrm{d}\mathbf{w}_t,
\end{equation}
where $(\eta, \gamma)$ are score and noise weight, $g(t)$ is the diffusion coefficient, for which we use $g(t) = \frac{1-t}{t}$, and the score is derived from the predicted velocity through the stochastic-interpolant identity~\citep{song2020score,lipman2022flow}
\begin{equation}
    s_\theta(\mathbf{x}_t,t,\hat{\mathbf{Z}})
    =
    \frac{tv_\theta(\mathbf{x}_t,t,\hat{\mathbf{Z}}) - \mathbf{x}_t}{(1-t)}.
\end{equation}

\section{Experiment}
\label{sec:experiment}
In this section, we evaluate whether \ours serves as a high-fidelity, versatile, and plug-and-play protein structure tokenizer by training and benchmarking it against several widely adopted baselines across multiple tasks. We use a filtered version of the Foldseek-clustered AFDB comprising 598,656 samples, and follow Kanzi in truncating sequences to 256 residues~\citep{barrio2023clustering,dilip2026flow}. The dataset contains no missing atoms and covers a broad range of structural distributions. During training, \ours converges after seeing 7.7M training samples, and we use the resulting checkpoint for all experiments. Further details are provided in App.~\ref{app:training_data_preprocessing}.

Our baselines cover representative protein structure tokenizer designs. ProTokens, IST, FoldToken4, DPLM2, and ESM3 are widely used learned discrete tokenizers for the protein modality, and have demonstrated strong adaptability to downstream modeling tasks~\citep{lin2023protokens,gao2024foldtoken4,gaujac2024learning,wang2025dplm,hayes2025simulating}. For FoldToken4, we use hierarchy level 12, corresponding to a vocabulary size of 4,096. We also include Kanzi, a recent flow-matching–based tokenizer, to evaluate the improvements achieved by \ours within the same technical paradigm~\citep{dilip2026flow}. We assess fidelity using the reconstruction task, and further consider unconditional generation, folding (conditional generation), and protein function understanding as representative downstream tasks. Finally, we conduct ablation studies on our design choices. Unless otherwise specified, we use 20 sampling steps for both the ODE and SDE samplers. The hyperparameters and search space for \ours are provided in App.~\ref{app:experimental-config}.

\subsection{Reconstruction}
\label{sec:reconstruction}

We evaluate \ours on a wide range of out-of-distribution test sets and show that it exhibits minimal bias toward the output distribution at inference time. This avoids degenerate behavior where the decoder relies on limited information from the quantizer and effectively performs a re-folding process. We argue that faithfully preserving the original structure is a necessary condition for applying protein structure tokenizers to downstream tasks.

Given that most protein structure tokenizers are trained on real monomer structures or synthetic data generated by folding models, we further assess \ours on structural fragments that cannot exist independently in biological systems (i.e., outside the distribution of folding model outputs), such as single chains extracted from multimeric complexes or motif fragments. These results further support that \ours exhibits minimal output distribution bias.

\begin{wrapfigure}{r}{0.3\linewidth}
    \centering
    \includegraphics[width=1\linewidth]{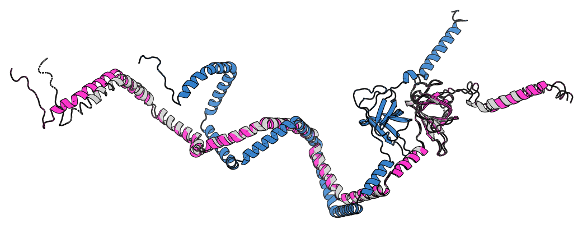}
    \caption{Reconstruction results of DPLM2 and \ours on T1137s3. \textcolor{gray}{Gray}: ground truth, \textcolor{purple}{purple}: \ourscomma, \textcolor{blue}{blue}: DPLM2}
    \label{fig:badcase}
\end{wrapfigure}

Tab.~\ref{tab:reconstruction} presents our results on four test sets, including CAMEO2022, CASP15, CASP16, and a held-out $D_{\text{FS}}$ subset~\citep{haas2018continuous,simpkin2023tertiary,jing2023eigenfold,kryshtafovych2026progress}. Reconstruction quality is evaluated using $\mathrm{C}_\alpha$ RMSD and TM-score~\citep{zhang2004scoring}. When using the SDE sampler, we set $\eta = 0.5$. Detailed metric definitions are provided in App.~\ref{sec:metric-definitions}.

\begin{table}[ht]
\centering
\caption{Reconstruction accuracy results. Subscripts denote standard deviations, with values in RMSD columns reported in units of $10^{-2}$ and values in TM-score columns reported in units of $10^{-3}$.}
\label{tab:reconstruction}
\small
\begin{tabular}{lcc|cc|cc|cc}
\toprule
 & \multicolumn{2}{c}{CAMEO2022} & \multicolumn{2}{c}{CASP15} & \multicolumn{2}{c}{CASP16} & \multicolumn{2}{c}{$D_{\text{FS}}$} \\
\cmidrule(r){2-3} \cmidrule(r){4-5} \cmidrule(r){6-7} \cmidrule(r){8-9}
 & RMSD $\downarrow$ & TM $\uparrow$ & RMSD $\downarrow$ & TM $\uparrow$ & RMSD $\downarrow$ & TM $\uparrow$ & RMSD $\downarrow$ & TM $\uparrow$ \\
\midrule
IST & 3.86 & .744 & 8.64 & .688 & 6.67 & .676 & 3.55 & .721 \\
DPLM2 & 1.87 & .903 & 8.52 & .792 & 8.30 & .767 & 2.07 & .872 \\
FoldToken4 & 1.22 & .941 & 6.47 & .845 & 2.55 & .871 & 1.40 & .911 \\
ESM3 & 1.02 & \uline{.969} & 1.99 & .918 & 3.41 & .891 & 0.91 & .962 \\
ProTokens & \uline{0.94} & .963 & \uline{1.23} & \uline{.965} & \uline{1.25} & \uline{.957} & 0.89 & .958 \\
\midrule
kanzi\tablefootnote{See App.~\ref{app:kanzi-length-generalization} for a discussion of the differences between our values and those reported in the Kanzi paper.} & 9.33 & .674 & 21.86 & .520 & 13.33 & .604 & 0.98 & .943 \\
\rowcolor{gray!20} \ourscomma (ODE) & 0.96$_{2.0}$ & .964$_{0.4}$ & 1.48$_{7.6}$ & .954$_{1.5}$ & 1.41$_{4.7}$ & .952$_{1.9}$ & \uline{0.71}$_{0.8}$ & \uline{.978}$_{0.5}$ \\
\rowcolor{gray!20} \ourscomma (SDE) & \textbf{0.64$_{0.5}$} & \textbf{.983$_{0.2}$} & \textbf{1.19$_{3.8}$} & \textbf{.967$_{0.8}$} & \textbf{1.01$_{2.0}$} & \textbf{.977$_{0.5}$} & \textbf{0.42$_{0.3}$} & \textbf{.988$_{0.1}$} \\
\bottomrule
\end{tabular}
\end{table}

\begin{figure}[t]
    \centering
    \includegraphics[width=0.8\linewidth]{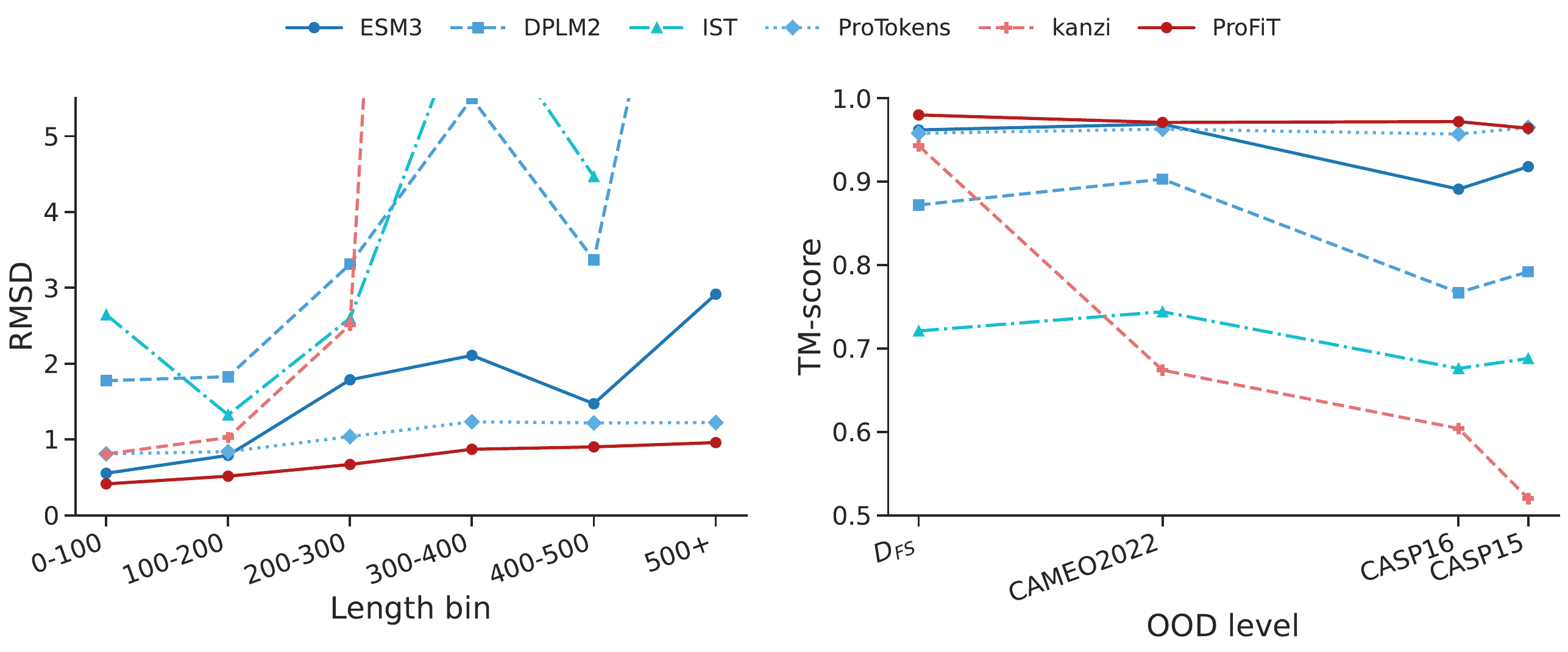}
    \caption{\ours exhibits stronger generalization, with reconstruction error largely insensitive to both protein length and the OOD level. The OOD level is quantified using pairwise TM-score against the PDB dataset, and is normalized such that 0 corresponds to in-distribution samples and 1 corresponds to the most out-of-distribution samples.}
    \label{fig:length_ood_vs_reconstruction}
\end{figure}

Tab. \ref{tab:reconstruction} shows that \ours consistently achieves either the best or near-best reconstruction performance across all evaluated settings, demonstrating strong and stable preservation of protein backbone structure. Despite a relatively compact model size, it remains competitive with larger baselines, indicating efficient utilization of model capacity.

We further observe that, as the distribution shift of the test sets increases, the test sets include structures that cannot exist as independently foldable monomers (e.g., T1137s3, see Fig.~\ref{fig:examples} \& \ref{fig:badcase}), the reconstruction error of \ours increases only marginally, and its output distribution does not collapse toward the distribution of monomeric proteins. For example, as shown in Fig.~\ref{fig:badcase}, the reconstruction produced by DPLM2 tends to “refold” the protein into a plausible monomeric structure rather than faithfully reconstructing the original input. Such robustness suggests that \ours does not overfit to the implicit output distributions of its training sources, but instead learns more generalizable structural representations.

\subsection{Unconditional Generation}
\label{sec:unconditional_generation}
In this section, we evaluate \ours on the unconditional generation task. To cover a range of practical scenarios, we assess all baselines and \ours under both autoregressive (AR) and discrete diffusion architectures. For the AR architecture, we train a 220M-parameter decoder-only Transformer to model protein tokens, using BOS and EOS tokens to indicate the beginning and end of each sequence. For the discrete diffusion architecture, we directly follow the official training script to train DPLM2. We generated 250 samples for each tokenizer under both the AR and DPLM architectures. For DPLM, following its original setting, we sampled 50 sequences at each length of 100, 200, 300, 400, and 500 residues. We evaluate generation performance in terms of quality and diversity. In the AR generation experiments, we set $\eta=0.5, \gamma=0$; in the DPLM2-650M generation experiments, we set $\eta=0.25, \gamma=0.25$ \citep{wang2025dplm}. Details of the AR model and detailed metric definitions are provided in App.~\ref{sec:metric-definitions}.

\begin{table}[t]
\centering
\caption{Results for unconditional generation with the AR and DPLM2 architectures. Des., Div., and Nov. denote designability, diversity, and novelty, respectively. Standard deviations from five independent runs are indicated in the lower-right corner. Because Kanzi does not provide a backbone-level checkpoint, we cannot evaluate its secondary-structure composition.}
\label{tab:unconditional_generation}
\small
\begin{tabular}{lccc|ccccc}
\toprule
& \multicolumn{3}{c}{Quality} & \multicolumn{5}{c}{Diversity} \\
\cmidrule(r){2-4} \cmidrule(r){5-9}
& scRMSD $\downarrow$ & scTM $\uparrow$ & Des. $\uparrow$ & Div. $\downarrow$ & Nov. $\downarrow$ & $\alpha$\% & $\beta$\% & c\% \\
\midrule
\multicolumn{9}{c}{\textbf{AR}} \\
\midrule
DPLM2 & 8.75\std{.60} & .626\std{.027} & .71\std{.03} & .348\std{.024} & .831\std{.031} & .411\std{.003} & .071\std{.009} & .518\std{.010} \\
FoldToken4 & 8.07\std{.64} & .672\std{.012} & .60\std{.08} & .232\std{.015} & .759\std{.012} & .451\std{.017} & .080\std{.003} & .469\std{.017} \\
ESM3 & 6.74\std{.16} & .665\std{.008} & .69\std{.01} & .241\std{.001} & .791\std{.044} & .358\std{.004} & .198\std{.005} & .444\std{.006} \\
IST & 6.36\std{.85} & .724\std{.021} & .80\std{.02} & .229\std{.020} & .773\std{.031} & .386\std{.013} & .200\std{.011} & .414\std{.010} \\
ProTokens & \textbf{5.19\std{.22}} & \textbf{.800\std{.008}} & \textbf{.84\std{.03}} & .223\std{.013} & .810\std{.011} & .335\std{.017} & .071\std{.005} & .595\std{.012} \\
\midrule
kanzi & 12.92\std{.61} & .421\std{.021} & .63\std{.05} & .289\std{.007} & .729\std{.010} & - & - & - \\
\rowcolor{gray!20} \ourscomma & \uline{5.80\std{.33}} & \uline{.730\std{.019}} & \uline{.83\std{.04}} & .230\std{.022} & .715\std{.012} & .453\std{.013} & .126\std{.007} & .421\std{.006} \\
\midrule
\multicolumn{9}{c}{\textbf{DPLM2}} \\
\midrule
DPLM2 & \textbf{2.54\std{.44}} & \textbf{.879\std{.004}} & \textbf{.98\std{.01}} & .211\std{.001} & .650\std{.015} & .487\std{.006} & .069\std{.007} & .444\std{.003} \\
FoldToken4 & 16.50\std{.97} & .660\std{.010} & .62\std{.01} & .408\std{.017} & .651\std{.046} & .178\std{.008} & .095\std{.004} & .727\std{.007} \\
ESM3 & 19.30\std{1.52} & .600\std{.013} & .67\std{.03} & .374\std{.041} & .515\std{.025} & .509\std{.008} & .109\std{.010} & .382\std{.003} \\
IST & 21.94\std{1.07} & .499\std{.018} & .35\std{.04} & .300\std{.003} & .658\std{.008} & .787\std{.019} & .059\std{.005} & .154\std{.017} \\
ProTokens & 7.42\std{.41} & .640\std{.003} & .74\std{.01} & .314\std{.012} & .829\std{.009} & .486\std{.002} & .008\std{.001} & .505\std{.002} \\
\midrule
kanzi & 11.64\std{.41} & .494\std{.004} & .45\std{.02} & .346\std{.008} & .701\std{.010} & - & - & - \\
\rowcolor{gray!20} \ourscomma & \uline{6.62\std{.20}} & \uline{.695\std{.002}} & \uline{.83\std{.02}} & .290\std{.008} & .649\std{.004} & .651\std{.015} & .061\std{.001} & .287\std{.015} \\
\bottomrule
\end{tabular}
\end{table}

Tab.~\ref{tab:unconditional_generation} shows that \ours is the only tokenizer that maintains consistently strong performance across both autoregressive and discrete diffusion architectures, without any architecture specific customization. In contrast, state-of-the-art methods tend to specialize in a single generative paradigm: ProTokens achieves its strongest results under the autoregressive setting but degrades under diffusion, while DPLM2 exhibits the opposite behavior. This suggests that existing tokenizers are often tightly coupled to the inductive biases of their respective architectures, whereas \ours provides a more generalizable representation that transfers robustly across both regimes.

\begin{wrapfigure}{r}{0.4\linewidth}
    \centering
    \includegraphics[width=1\linewidth]{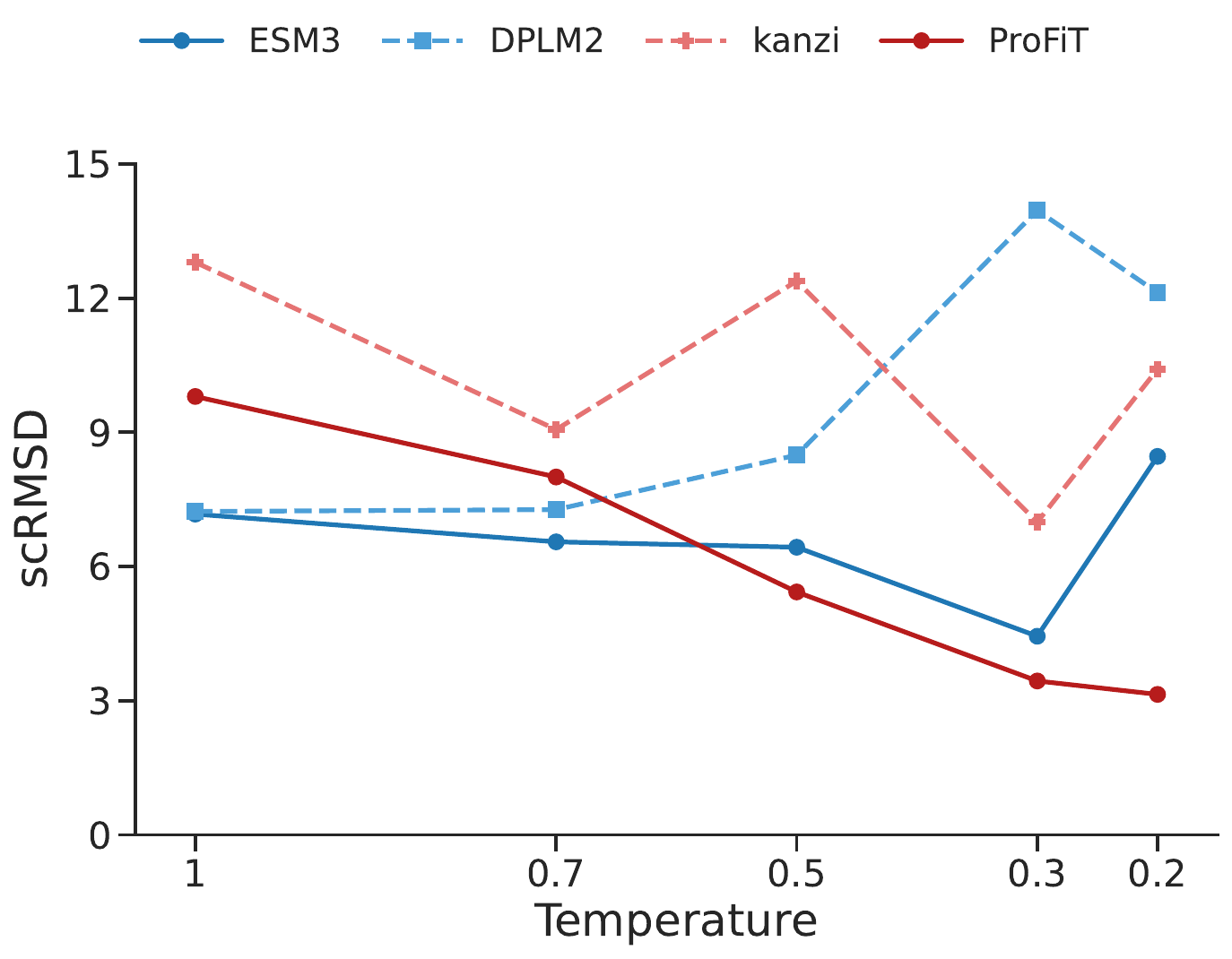}
    \caption{Effect of sampling temperature on AR generation quality.}
    \label{fig:temp_vs_scrmsd}
    \vspace{-20pt}
\end{wrapfigure}

In Fig.~\ref{fig:temp_vs_scrmsd}, we additionally provide a comparison between autoregressive sampling temperature and generation quality. When modeling \ours tokens with an autoregressive model, the AR likelihood serves as a strong indicator of the quality of the generated structures. This suggests that \ours enables the likelihood estimated by the AR model to be used as a proxy for early screening of generation quality.

\subsection{Conditional Generation}
\label{sec:conditional_generation}
To evaluate the performance of different tokenizers in cross-modal tasks, we select the representative folding task as our conditional generation benchmark. Given the strong performance of DPLM2 on folding in the discrete domain, we retrain DPLM2 using the same training dataset but with different structure tokenizers for discretization, thereby assessing the adaptability of \ours to multimodal data~\citep{wang2025dplm}. We evaluate on CAMEO2022 using the same metrics as in Sec. \ref{sec:reconstruction}. In each run, we generate one structure per sequence and repeat the evaluation over five independent runs.

Overall, \ours achieves strong performance in the conditional folding setting, reaching results that are competitive with tokenizers specifically optimized for DPLM2. This indicates that the learned tokenizer generalizes well to cross-modal conditional generation tasks and retains strong structural fidelity even in challenging folding scenarios.

\subsection{Understanding}
We evaluate \ours on protein understanding tasks in Tab.~\ref{tab:understanding}. Following the setup of StructTokenBench, we select a subset of its functional site prediction tasks \cite{yuan2025protein}. Functional site prediction is a residue-level binary classification task that determines whether each residue belongs to a particular type of functional site. We report AUROC for residue-level binary classification across Fold and SupFam splits. Notably, \ours is trained without any auxiliary losses, meaning that we do not explicitly enforce semantic alignment of the tokens. Nevertheless, \ours naturally acquires semantically meaningful structural representations.

\begin{table}[t]
\centering

\begin{minipage}[t]{0.34\linewidth}
\vspace{0pt}
\centering
\caption{Discrete diffusion conditional generation.}
\label{tab:folding}
\small
\begin{tabular}{lcc}
\toprule
 & \multicolumn{2}{c}{CAMEO2022} \\
\cmidrule(r){2-3}
 & RMSD $\downarrow$ & TM $\uparrow$ \\
\midrule
IST & 10.69\std{1.03} & .516\std{.032} \\
ProTokens & 9.84\std{1.10} & .548\std{.035} \\
FoldToken4 & 9.05\std{1.40} & .639\std{.035} \\
ESM3 & 8.75\std{1.19} & .649\std{.035} \\
DPLM2 & \textbf{8.06}\std{1.04} & \textbf{.765}\std{.035} \\
\midrule
kanzi & 16.42\std{1.62} & .320\std{.031} \\
\rowcolor{gray!20}
\ourscomma & \uline{8.61}\std{1.27} & \uline{.724}\std{.036} \\
\bottomrule
\end{tabular}
\end{minipage}
\hfill
\begin{minipage}[t]{0.63\linewidth}
\vspace{0pt}
\centering
\caption{Functional site prediction performance on StructTokenBench.}
\label{tab:understanding}
\small
\begin{tabular}{l l c c c >{\columncolor{gray!20}}c}
\toprule
\textbf{Task} & \textbf{Split}
& \multicolumn{4}{c}{\textbf{Model}} \\
\cmidrule(lr){3-6}
& & Foldseek & ProTokens & ESM3 & \ourscomma \\
\midrule
\multirow{2}{*}{BindInt}
& Fold   & \textbf{53.18} & 44.66 & 44.30 & \uline{44.97} \\
& SupFam & 46.20 & \uline{86.05} & \textbf{90.77} & 81.44 \\
\multirow{2}{*}{Con}
& Fold   & 49.26 & \uline{56.23} & 55.22 & \textbf{56.94} \\
& SupFam & 51.39 & 74.33 & \textbf{80.53} & \uline{79.19} \\
\multirow{2}{*}{Rep}
& Fold   & 47.70 & \textbf{77.25} & 74.70 & \uline{75.30} \\
& SupFam & 52.53 & 78.90 & \textbf{82.36} & \uline{79.42} \\
\multirow{2}{*}{Ept}
& Fold   & 54.52 & 54.69 & \textbf{63.69} & \uline{55.11} \\
& SupFam & 50.56 & \textbf{67.52} & 61.97 & \uline{63.31} \\
\bottomrule
\end{tabular}
\end{minipage}

\end{table}

\subsection{Codebook Utilization}

\begin{wraptable}{r}{0.48\textwidth}
    \vspace{-12pt}
    \centering
    \caption{Codebook utilization across all baselines.}
    \label{tab:codebook_utilization}
    \small
    \begin{tabular}{lccc}
        \toprule
        Model & UR $\uparrow$ & PPL/K $\uparrow$ & Similarity $\downarrow$ \\
        \midrule
        \rowcolor{gray!20}
        \ourscomma & \textbf{98.91\%} & \textbf{86.83\%} & \uline{0.959} \\
        ProTokens  & \uline{98.24\%} & 66.15\% & 0.994 \\
        ESM3       & 92.36\% & \uline{74.70\%} & 0.984 \\
        DPLM2      & 82.76\% & 63.48\% & \textbf{0.846} \\
        Kanzi      & 71.20\% & 59.78\% & 0.986 \\
        FoldToken4 & 52.93\% & 19.09\% & 0.987 \\
        IST        & 27.88\% & 22.31\% & 0.961 \\
        \bottomrule
    \end{tabular}
    \vspace{-10pt}
\end{wraptable}

We assess codebook utilization on CAMEO2022 dataset, using the utilization rate, normalized perplexity, and code similarity. As shown in Tab.~\ref{tab:codebook_utilization}, \ours achieves the highest UR and PPL/K despite having the largest vocabulary and a low-dimensional codebook. It also has the second-lowest code similarity, surpassed only by DPLM2, which uses LFQ~\citep{yu2024language,wang2025dplm}. These results indicate broad and relatively balanced use of \ours's codebook, consistent with the intended effect of dead-code replacement.

\section{Discussions}
\label{sec:discussions}
In this work, we present \ourscomma, a lightweight flow-matching-based protein structure tokenizer with a compact discrete bottleneck and a conditional continuous-time decoder. By combining a simple objective with stable VQ training, \ours establishes a new state of the art among flow-matching-based protein structure tokenizers while achieving strong reconstruction fidelity, transferability, and emergent token semantics without auxiliary semantic losses. These results suggest that effective protein tokenization does not necessarily require heavily engineered multi-term objectives or large equivariant backbones, and highlight several practical directions for future flow-matching tokenizers: robust codebook dynamics, evaluation beyond standard in-distribution settings, and improved path design and conditioning strategies for better scalability and robustness across biological modalities.

Flow-matching–based tokenization is still far from fully explored. Important open questions include how such tokenizers scale, and whether other generative modeling paradigms may be better suited for protein modeling. These remain open questions for future work.

\subsection*{AI use statement}
In this work, we used generative AI tools to implement methods, assist with translation, clean and reformat datasets, support qualitative and thematic analysis, interpret results, and create scientific figures and images. We did not use generative AI tools to generate synthetic datasets, develop theoretical models or conceptual frameworks, or design or provide feedback on research methodologies or experiments. The use of generative AI tools to formulate mathematical claims, provide critical ingredients for proving mathematical claims, assist in writing proofs, or propose or refine hypotheses was not applicable to this work. We reviewed all code and suggestions generated by LLMs and verified their correctness. We take full responsibility for the final content of this work, including all text, claims, and artifacts produced with the assistance of generative AI.

\subsection*{Ethics statement}
This work does not involve human or animal subjects, personal or sensitive information, or interactions with individuals. We use publicly available protein structure datasets and comply with their stated terms of use. The methods and analyses presented here do not introduce identifiable privacy or security risks, and we are not aware of any foreseeable harmful applications or discrimination-related concerns arising from this work. We have no conflicts of interest or external sponsorship to declare. To the best of our knowledge, this study complies with applicable research-integrity and legal requirements.

\subsection*{Reproducibility statement}
We describe all methodological design choices in Sec.~\ref{sec:method} and provide comprehensive training and experimental details, including model hyperparameters, in App.~\ref{app:experiment-details} to facilitate reproducibility. We also provide an anonymous repository for convenient reproduction and will release the datasets and full repository in a subsequent public release.



\bibliography{iclr2027_conference}
\bibliographystyle{iclr2027_conference}

\newpage
\appendix
\section{Metric Definitions}
\subsection{Overall Score Definitions}
\label{app:overall_score_definitions}
\begin{figure}[ht]
    \centering
    \includegraphics[width=0.75\linewidth]{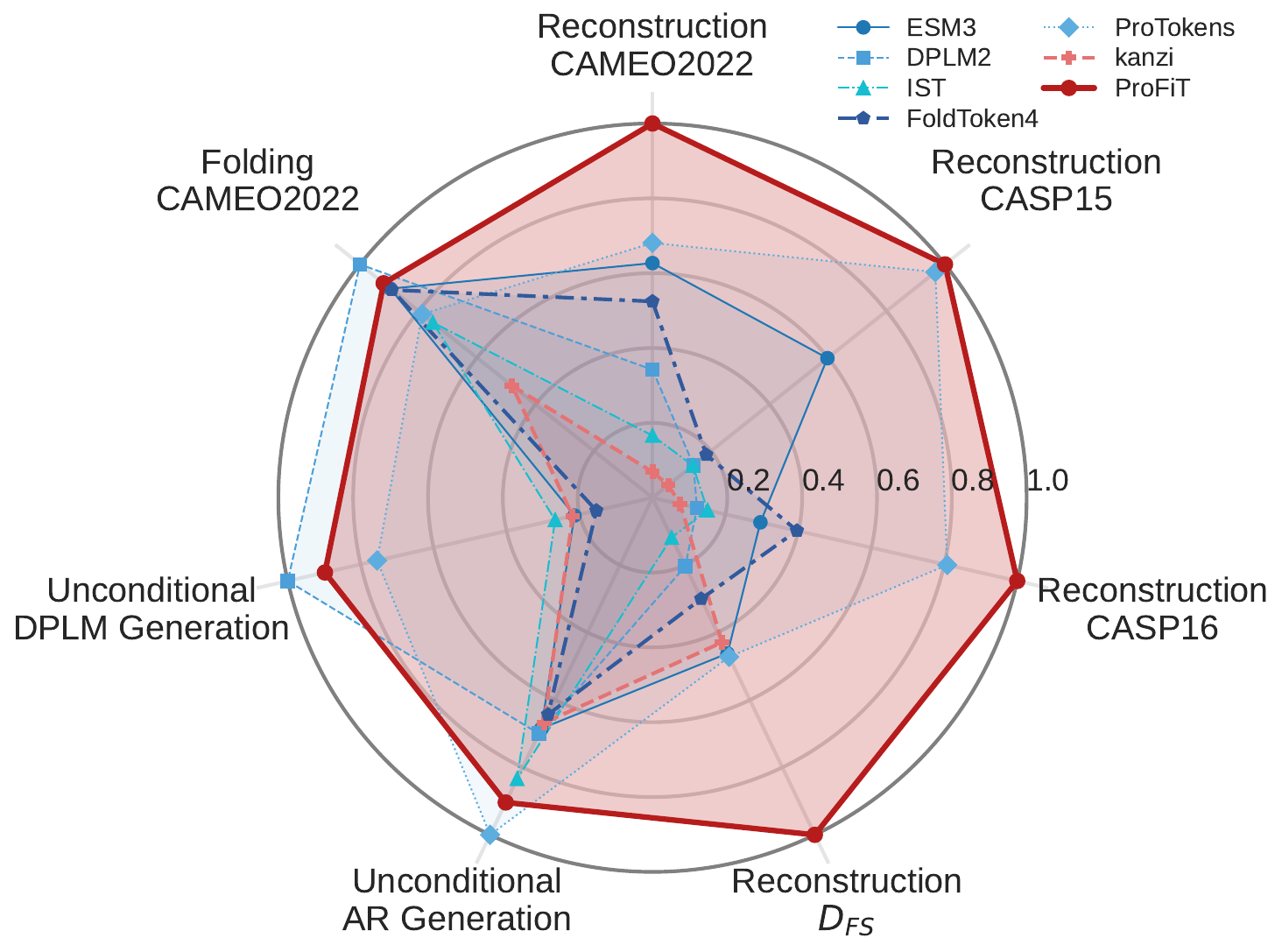}
    \caption{Relative performance comparison.}
    \label{fig:radar}
\end{figure}

The overall scores in Fig.~\ref{fig:main} are derived from the relative performance comparison between \ours and other tokenizers, where all metrics are reported in RMSD or scRMSD. Each score is normalized by the best-performing method for the corresponding metric, such that the highest score is scaled to 1. The overall reconstruction score is computed from the results on the four reconstruction datasets, while the overall generation score is computed from the remaining three generation-related evaluations. More detailed score comparisons are provided in Fig.~\ref{fig:radar}.

\subsection{Experiment Metric Definitions}
\label{sec:metric-definitions}

\paragraph{Root-Mean-Square Deviation (RMSD).}
RMSD is a widely used metric for evaluating the geometric similarity between predicted and native protein structures. For a protein of $L$ residues, given predicted coordinates ($\{x_i\}_{i=1}^L$) and reference coordinates ($\{y_i\}_{i=1}^L$), RMSD is computed as:
\begin{equation}
\text{RMSD} = \sqrt{\frac{1}{L} \sum_{i=1}^L | x_i - y_i |^2}.
\end{equation}
Before computation, the predicted structure is rigidly aligned to the reference structure to remove the effects of global translation and rotation, using the Kabsch alignment algorithm. RMSD is sensitive to local deviations and provides a straightforward measure of absolute structural accuracy.

\paragraph{Template Modeling Score (TM-Score).}
TM-Score measures the global structural similarity between two protein structures while normalizing for protein length, providing a more robust metric for evaluating fold-level correctness. It is defined as:
\begin{equation}
\text{TM-Score} = \max \left[ \frac{1}{L_{\text{ref}}} \sum_{i=1}^{L_{\text{ali}}} \frac{1}{1 + (d_i/d_0(L_{\text{ref}}))^2} \right].
\end{equation}
where $d_i$ is the distance between aligned residues, $L_{\text{ref}}$ is the length of the reference protein, $L_{\text{ali}}$ is the number of aligned residues, and $d_0$ is a length-dependent normalization factor. TM-Score values range from 0 to 1, with higher scores indicating better overall fold similarity, and are less sensitive to local structural noise compared to RMSD.

\paragraph{scRMSD and scTM.}
We evaluate self-consistency through a structure-to-sequence-to-structure loop. For each generated structure token sequence, we first decode a 3D backbone, then design amino-acid sequences with ProteinMPNN (8 sequences per target, sampling temperature 0.1, $\mathrm{C}_\alpha$-only mode), and finally refold all designed sequences with ESMFold~\citep{dauparas2022robust,lin2023evolutionary}. We report the best-case values across refolded candidates: $\mathrm{scRMSD}=\min$ RMSD and $\mathrm{scTM}=\max$ TM-score. RMSD is computed on $\mathrm{C}_\alpha$ atoms after Kabsch alignment, and TM-score is computed with the TMscore binary on $\mathrm{C}_\alpha$-only PDBs.

\paragraph{Designability (Des.).}
Designability is defined as the proportion of samples with $\mathrm{scTM}>0.5$.

\paragraph{Diversity (Div.).}
Diversity is computed only among designable structures. We define a valid pair as a pair of designable proteins whose length difference does not exceed 10. The reported diversity is the mean pairwise TM-score across valid pairs.

\paragraph{Novelty (Nov.).}
Novelty is evaluated for designable samples by computing their maximum TM-score against the PDB dataset, and the reported overall Novelty is the mean of these per-sample maxima.

\paragraph{Secondary Structure Composition ($\alpha/\beta/c$).}
Secondary structure is assigned from decoded structures using DSSP on backbone atoms~\citep{kabsch1983dictionary}. We parse DSSP labels and map $H/G/I\rightarrow\alpha$-helix, $E/B\rightarrow\beta$-strand, and all others to coil.

\paragraph{Utilization Rate (UR).}
UR measures the fraction of codebook entries that are used at least once over the evaluation set. Given a codebook of size $K$ and token counts $\{n_k\}_{k=1}^{K}$, we compute
\[
\mathrm{UR}=\frac{1}{K}\sum_{k=1}^{K}\mathbb{I}(n_k>0).
\]

\paragraph{Normalized Perplexity (PPL/K).}
Let $p_k=n_k/\sum_{j=1}^{K}n_j$ denote the empirical frequency of code $k$. We first compute the code usage perplexity as
\[
\mathrm{PPL}=\exp\left(-\sum_{k=1}^{K}p_k\log p_k\right),
\]
and report its value normalized by the codebook size, $\mathrm{PPL}/K$. A value closer to one indicates more balanced code usage.

\paragraph{Codebook Similarity.}
To measure redundancy among codebook entries, we compute the maximum cosine similarity between each codebook vector and any other distinct codebook vector, and then average these values over all entries. Lower similarity indicates a more diverse codebook with less redundancy.

LFQ does not maintain an explicit embedding matrix. Instead, each code corresponds to a binary vector in the implicit codebook $\mathcal{C}_{\mathrm{LFQ}}\subseteq\{-1,+1\}^{d}$~\citep{yu2024language}. We reconstruct these binary vectors from the code indices and compute codebook similarity using the same nearest-neighbor cosine-similarity metric:
\[
\mathrm{Sim}(\mathcal{C}_{\mathrm{LFQ}})=
\frac{1}{|\mathcal{C}_{\mathrm{LFQ}}|}
\sum_{\mathbf{c}\in\mathcal{C}_{\mathrm{LFQ}}}
\max_{\mathbf{c}'\neq\mathbf{c}}
\frac{\mathbf{c}^{\top}\mathbf{c}'}
{\|\mathbf{c}\|_2\|\mathbf{c}'\|_2}.
\]

\section{Experiment Details}
\label{app:experiment-details}
\subsection{Training Data Preprocessing}
\label{app:training_data_preprocessing}
All training samples are drawn from the representative sequences of each cluster in the AFDB Foldseek cluster dataset. We retain only sequences with length $\leq 256$ and pLDDT $\geq 75$. In addition, all homologous sequences overlapping with the test sets are removed, using a Foldseek structural similarity threshold of 80\%.

Additional data augmentation during training includes centering each structure at the origin and applying random rotations. We do not perform any additional finetuning on longer sequences.

\subsection{Experimental Configuration}
\label{app:experimental-config}
For the \ours checkpoint used in our experiments, we train the model on 8 H20 GPUs with a per-GPU batch size of 8, resulting in a total effective batch size of 64. Training is conducted for 120k steps and takes approximately 20 hours in wall-clock time. Tab.~\ref{tab:experimental-config} summarizes the hyperparameter search spaces used during training. For hyperparameters that were not extensively tuned, we report the values used in our experiments directly. For hyperparameters explored through grid search, we use \textbf{boldface} to indicate the final selected values.

\begin{table}[ht]
\centering
\caption{Hyperparameters for \ourscomma.}
\label{tab:experimental-config}
\begin{tabular}{ll}
\toprule
Hyperparameter & Value \\
\midrule
Train/validation split & $0.999/0.001$ \\
Encoder layers & 5 \\
Decoder layers & 5 \\
Attention heads & 8 \\
Token dimension & 256 \\
Latent dimension, $d$ & 8 \\
Codebook size, $K$ & 8192 \\
VQ EMA decay & 0.99 \\
Dead-code threshold, $\tau_{\mathrm{dead}}$ & 0.5 / 1 / \textbf{2} / 4 \\
Dead-code reset mass, $\rho$ & 5 / 10 / \textbf{20} / 40 \\
Maximum token masking ratio, $r$ & \textbf{0.1} / 0.2 / 0.4 \\
Optimizer & Adam(0.9, 0.999) \\
Learning rate & 5e-3 / 1e-3 / 5e-4 / \textbf{1e-4} / 5e-5 \\
Precision & bf16 mixed \\
Model EMA decay & 0.999 \\
\bottomrule
\end{tabular}
\end{table}

\subsection{Details of the AR Model}
For the unconditional generation experiments in Sec.~\ref{sec:unconditional_generation}, we train a Transformer model for autoregressive modeling of protein structure tokens. BOS and EOS tokens are appended to the beginning and end of each sequence, respectively, and PAD tokens are used for padding within a batch. The Transformer consists of 16 self-attention blocks with a hidden dimension of 1280 and 16 attention heads, using Pre-LayerNorm and GELU activations. The FFN expansion ratio is set to 2, with a dropout rate of 0.1. The input token embedding and output projection head share weights.

Training is performed on exactly the same dataset used for tokenizer training. We use the Adam optimizer with a learning rate of $3\times10^{-4}$, betas $(0.9, 0.98)$, weight decay $0.01$, and 2000 warmup steps. The model is trained for a total of 1k epochs.

\section{Extra Experiment Result}
\subsection{Analysis of Kanzi's Length Generalization}
\label{app:kanzi-length-generalization}
To investigate the discrepancy between our Kanzi results and those reported in the original paper, we evaluate Kanzi separately on proteins within and beyond its training truncation length of 256 residues. The results are summarized in Tab.~\ref{tab:kanzi-length-generalization}.

\begin{table}[ht]
    \centering
    \caption{Length-stratified performance of Kanzi. Each entry reports RMSD (\AA) / TM-score.}
    \label{tab:kanzi-length-generalization}
    \begin{tabular}{lccc}
        \toprule
        Length regime & CAMEO & CASP15 & CASP16 \\
        \midrule
        $\leq 256$ residues & 0.993 / 0.940 & 0.932 / 0.959 & 1.341 / 0.939 \\
        $>256$ residues & 20.63 / 0.320 & 36.06 / 0.223 & 22.91 / 0.343 \\
        \bottomrule
    \end{tabular}
\end{table}

Kanzi performs well on proteins within the training truncation length but degrades sharply on longer proteins, indicating limited length generalization. This observation provides a plausible explanation for the discrepancy between our results and those reported in the original Kanzi paper. Specifically, although we use the correct model weights, the preprocessing applied to the test sets may differ from that used in the original evaluation. The original paper does not provide sufficient details to determine the exact preprocessing procedure. When we truncate the test proteins to 256 residues, we recover performance consistent with the original report, further supporting this explanation.

The same truncation length is used during the training of \ours. In contrast, \ours maintains nearly unchanged reconstruction quality on the untruncated test sets, as shown in Fig.~\ref{fig:length_ood_vs_reconstruction}. These results suggest that \ours generalizes substantially better beyond its training length.

\subsection{Out-of-Distribution Shift Across Evaluation Sets}
Fig.~\ref{fig:length_ood_vs_reconstruction} (right) shows that the reconstruction quality of several baselines degrades substantially as the evaluation sets shift further from the PDB distribution. To contextualize this trend, we include Protenix v1 as a folding-model baseline in Tab.~\ref{tab:protenix_ood}, using MSA input and a single generated sample~\citep{protenix2026protenix}. Protenix exhibits a similar decline with increasing OOD severity, whereas the reconstruction quality of \ours remains relatively stable across evaluation sets. This contrast highlights the different behavior of a folding model's structural prior and \ours's token-conditioned reconstruction objective on OOD structures.

\begin{table}[ht]
    \centering
    \caption{Mean TM-score of Protenix v1 across evaluation sets.}
    \small
    \label{tab:protenix_ood}
    \begin{tabular}{lcccc}
        \toprule
        Metric & $D_{\text{FS}}$ & CAMEO & CASP16 & CASP15 \\
        \midrule
        Mean TM-score & 0.9531 & 0.8609 & 0.7559 & 0.7502 \\
        \bottomrule
    \end{tabular}
\end{table}

\section{Ablation Study}
\subsection{Ablation on Design Choices}
We conduct ablation studies on the design choices in Sec.~\ref{sec:enhancements_VQ}, demonstrating that they play an important role in both reconstruction quality and semantic compression. These components jointly push the Pareto frontier of reconstruction and generation, rather than merely trading off between the two. As shown in Tab.~\ref{tab:ablation}, both components positively impact reconstruction and generation quality. 

\begin{figure}[h]
    \centering
    \includegraphics[width=\linewidth]{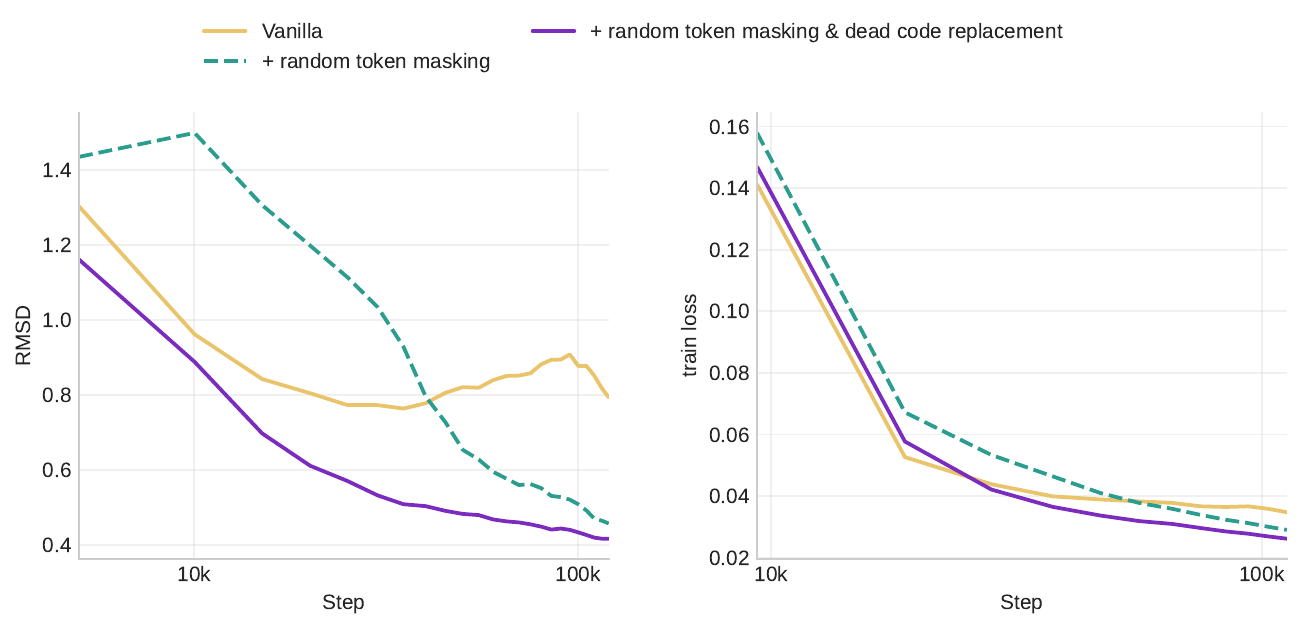}
    \caption{Ablation studies for Sec.~\ref{sec:enhancements_VQ}.}
    \label{fig:ablation}
\end{figure}

We further illustrate in Fig. \ref{fig:ablation} how our design choices alter the training dynamics. Both random token masking and dead code replacement promote better convergence. Although random token masking slightly increases the difficulty at the early stage of training, it creates a smoother loss landscape for optimization. We observe that without masking, the training loss can still decrease normally, but the reconstruction accuracy fluctuates severely and fails to converge to the optimum; with masking, the model exhibits stable and reproducible convergence behavior. Dead code replacement further strengthens this trend.

In addition, we find that the codebook usage of \ours during training follows a highly skewed long-tail distribution, which is unfavorable for downstream generation tasks. Dead code replacement effectively smooths the utilization across codes while still preserving an overall Zipf’s law pattern.

\begin{table}
\centering
\caption{Ablation study on CAMEO2022. We report reconstruction RMSD and AR generation scRMSD.}
\small
\label{tab:ablation}
\begin{tabular}{lcc}
\toprule
Method & Recon. RMSD $\downarrow$ & AR scRMSD $\downarrow$ \\
\midrule
\textbf{\ours} & \textbf{0.84} & \textbf{5.43} \\
\quad - token masking & 1.42 & 5.97 \\
\quad - code replacement & 1.45 & 6.55 \\
\bottomrule
\end{tabular}
\end{table}

\subsection{Ablation on Sampling Steps}
\begin{figure}[h]
    \centering
    \includegraphics[width=0.4\linewidth]{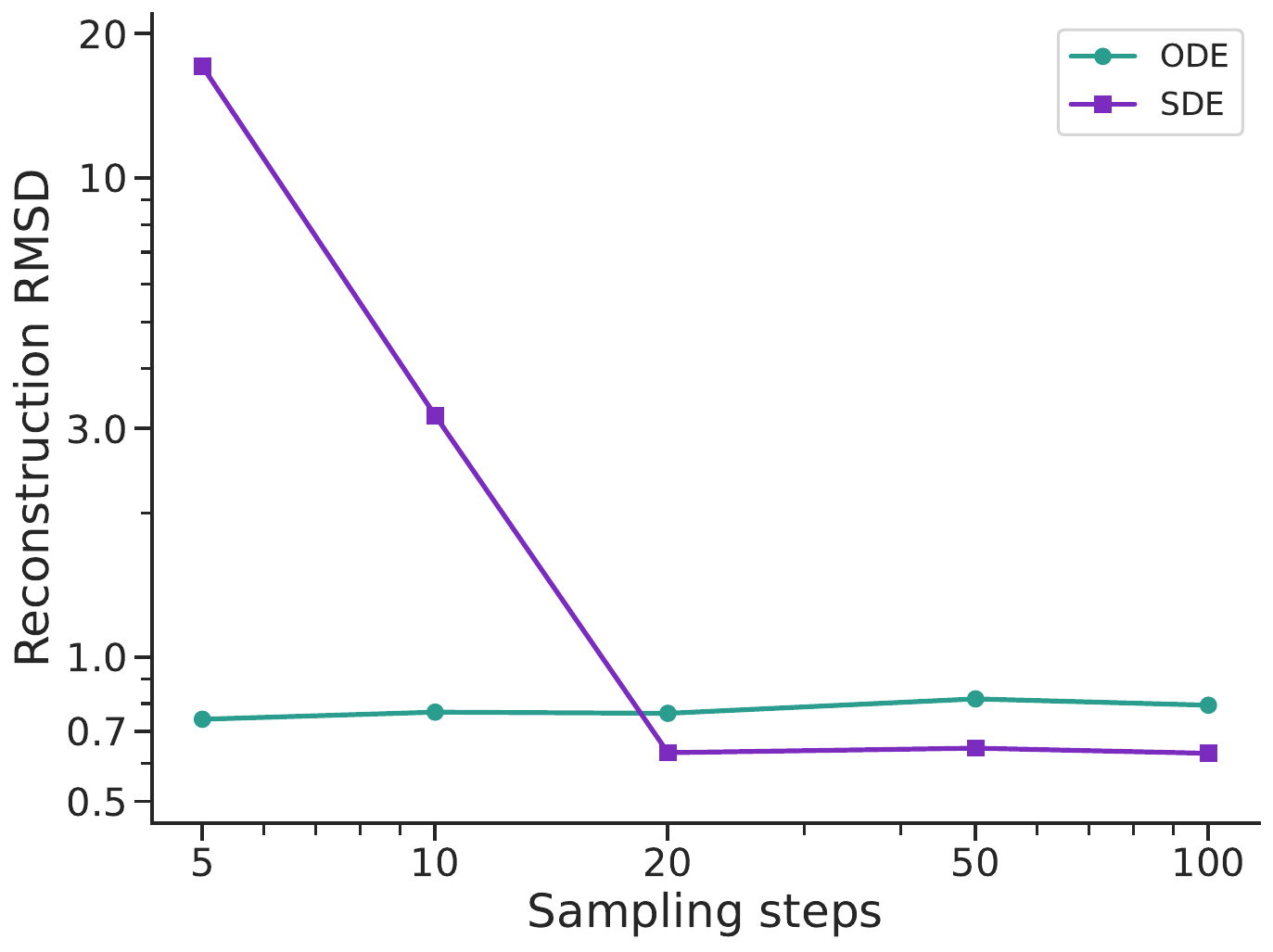}
    \caption{Performance under different numbers of sampling steps.}
    \label{fig:sampling_steps}
\end{figure}

We examine in Fig.~\ref{fig:sampling_steps} whether increasing the number of sampling steps at test time benefits the quality of the output structures. When using ODE sampler, \ours follows a straight sampling trajectory, and thus the output quality changes little as the number of sampling steps increases. When using SDE sampler ($\gamma = 0.5$), the output quality improves significantly with more sampling steps and already surpasses the ODE sampler at 20 steps. This is also the setting adopted in Sec.~\ref{sec:reconstruction}.

We furthur ablate the number of sampling steps in Tab.~\ref{tab:ode_steps} under ODE sampling. \ours achieves strong reconstruction performance with as few as 5 steps. Increasing the step count to 20 yields a substantial improvement primarily on CASP15, the most strongly OOD test set. This trend can also be observed in Fig.~\ref{fig:sampling_steps}, where the blue curve exhibits a slight upward trend. 

The sampling-step ablation indicates that \ours achieves high reconstruction fidelity without relying on long sampling trajectories, although additional steps can improve performance on the out-of-distribution CASP15 set. This behavior reflects the role of its stochastic decoder: conditioned on the encoder's tokens, the decoder is trained to recover the input structure rather than to generate a structure from an unconditional protein prior. The strong reconstruction results suggest that the tokens retain sufficient information for this recovery. Moreover, the simple flow-matching objective requires few task-specific structural assumptions, which may help limit bias toward particular structural preferences or evaluation metrics. In this setting, stochasticity and additional sampling steps support accurate conditional reconstruction; they do not necessarily imply a separate refolding stage.

\begin{table}[ht]
    \centering
    \caption{Effect of the number of ODE sampling steps on reconstruction performance. Each entry reports RMSD (\AA) / TM-score.}
    \small
    \label{tab:ode_steps}
    \begin{tabular}{ccccc}
        \toprule
        Steps & CAMEO & CASP15 & CASP16 & $D_{\text{FS}}$ \\
        \midrule
        2  & 0.76 / 0.977 & 4.31 / 0.897 & 1.78 / 0.953 & 0.67 / 0.977 \\
        5  & \textbf{0.68} / \textbf{0.981} & 2.73 / 0.924 & \textbf{1.34} / \textbf{0.968} & \textbf{0.49} / \textbf{0.985} \\
        20 & 0.96 / 0.964 & \textbf{1.48} / \textbf{0.954} & 1.41 / 0.952 & 0.71 / 0.978 \\
        \bottomrule
    \end{tabular}
\end{table}

\subsection{Ablation on Codebook Size}
We ablate the effect of codebook size by retraining \ours with $K=1000$ while keeping all other training and evaluation settings fixed, thereby matching Kanzi's vocabulary size~\citep{dilip2026flow}. We repeat the experiments from Sec.~\ref{sec:reconstruction}, \ref{sec:unconditional_generation}, and \ref{sec:conditional_generation}, and observe that the vocabulary size has no significant effect on the performance of \ourscomma. The results are presented in Tab.~\ref{tab:vocab-ablation-reconstruction}, \ref{tab:vocab-ablation-ar}, \ref{tab:vocab-ablation-dplm}, and \ref{tab:vocab-ablation-folding}.

\begin{table}[ht]
    \centering
    \caption{Codebook-size ablation for reconstruction. Each entry reports RMSD (\AA) / TM-score.}
    \label{tab:vocab-ablation-reconstruction}
    \begin{tabular}{lcccc}
        \toprule
        Method & CAMEO & CASP15 & CASP16 & $D_{\text{FS}}$ \\
        \midrule
        Kanzi & 9.33 / 0.674 & 21.86 / 0.520 & 13.33 / 0.604 & 0.98 / 0.943 \\
        \ours\ ($K=1000$) & 0.78 / 0.978 & 1.38 / 0.964 & 1.80 / 0.969 & 0.50 / 0.984 \\
        \ours\ ($K=8192$) & \textbf{0.64} / \textbf{0.983} &
        \textbf{1.19} / \textbf{0.967} &
        \textbf{1.01} / \textbf{0.977} &
        \textbf{0.42} / \textbf{0.988} \\
        \bottomrule
    \end{tabular}
\end{table}

\begin{table}[ht]
    \centering
    \caption{Codebook-size ablation for unconditional generation with the AR architecture.}
    \label{tab:vocab-ablation-ar}
    \begin{tabular}{lcccccccc}
        \toprule
        Method & Des. $\uparrow$ & scRMSD $\downarrow$ & scTM $\uparrow$ &
        Div. $\downarrow$ & Nov. $\downarrow$ & $\alpha$\% & $\beta$\% & c\% \\
        \midrule
        Kanzi & 0.63 & 12.38 & 0.429 & 0.289 & 0.721 & -- & -- & -- \\
        \ours\ ($K=1000$) & \textbf{0.82} & 5.80 & 0.710 & 0.251 & 0.712 & 0.457 & 0.115 & 0.428 \\
        \ours\ ($K=8192$) & \textbf{0.82} & \textbf{5.43} & \textbf{0.729} & 0.262 & 0.714 & 0.466 & 0.106 & 0.428 \\
        \bottomrule
    \end{tabular}
\end{table}

\begin{table}[ht]
    \centering
    \caption{Codebook-size ablation for unconditional generation with the DPLM architecture.}
    \label{tab:vocab-ablation-dplm}
    \begin{tabular}{lcccccccc}
        \toprule
        Method & Des. $\uparrow$ & scRMSD $\downarrow$ & scTM $\uparrow$ &
        Div. $\downarrow$ & Nov. $\downarrow$ & $\alpha$\% & $\beta$\% & c\% \\
        \midrule
        Kanzi & 0.48 & 11.35 & 0.496 & 0.346 & 0.706 & -- & -- & -- \\
        \ours\ ($K=1000$) & 0.79 & 6.74 & 0.660 & 0.299 & 0.654 & 0.621 & 0.072 & 0.307 \\
        \ours\ ($K=8192$) & \textbf{0.81} & \textbf{6.25} & \textbf{0.685} & 0.298 & 0.655 & 0.617 & 0.067 & 0.316 \\
        \bottomrule
    \end{tabular}
\end{table}

\begin{table}[ht]
    \centering
    \caption{Codebook-size ablation for conditional folding.}
    \label{tab:vocab-ablation-folding}
    \begin{tabular}{lcc}
        \toprule
        Method & RMSD (\AA) $\downarrow$ & TM-score $\uparrow$ \\
        \midrule
        Kanzi & 17.01 & 0.351 \\
        \ours\ ($K=1000$) & 8.96 & 0.750 \\
        \ours\ ($K=8192$) & \textbf{8.31} & \textbf{0.787} \\
        \bottomrule
    \end{tabular}
\end{table}

\subsection{Analysis of Computational Cost}
We compare the computational costs of Kanzi and \ours for encoding and structure reconstruction. The results are summarized in Tab.~\ref{tab:computational-cost}. Although \ours uses a larger vocabulary, its reconstruction cost is substantially lower than that of Kanzi.The encoding cost of \ours is higher, reflecting differences in parameter count and encoder architecture; however, the decoder dominates the cost of multi-step reconstruction, where \ours is considerably more efficient.

\begin{table}[ht]
    \centering
    \caption{Computational cost of Kanzi and \ourscomma.}
    \label{tab:computational-cost}
    \begin{tabular}{lccc}
        \toprule
        Model & Encoding & 1-step reconstruction & 20-step reconstruction \\
        \midrule
        \ours & 10.93 GFLOPs & \textbf{42.07 GFLOPs} & \textbf{0.633 TFLOPs} \\
        Kanzi & \textbf{2.98 GFLOPs} & 183.5 GFLOPs & 2.928 TFLOPs \\
        \bottomrule
    \end{tabular}
\end{table}

\end{document}